\documentclass[runningheads]{llncs}
\usepackage{graphicx}
\usepackage{amsmath,amssymb} 
\usepackage{color}

\usepackage[width=122mm,left=12mm,paperwidth=146mm,height=193mm,top=12mm,paperheight=217mm]{geometry}

\begin{document}
\pagestyle{headings}
\mainmatter
\def\ECCV18SubNumber{***}  

\title{Does Haze Removal Help CNN-based Image Classification?} 

\titlerunning{Does Haze Removal Help CNN-based Image Classification?}

%
\author{Yanting Pei\inst{1,2} \and
Yaping Huang\inst{1,}\thanks{Co-corresponding authors.} \and
Qi Zou\inst{1}  \and
Yuhang Lu\inst{2} \and
Song Wang\inst{2,3,\star}
}
  
%
\authorrunning{Y. Pei, Y. Huang, Q. Zou, Y. Lu, and S. Wang}
%

\institute{Beijing Key Laboratory of Traffic Data Analysis and Mining, Beijing Jiaotong University, Beijing, China \and
Department of Computer Science and Engineering, University of South Carolina, Columbia, SC, USA \and
School of Computer Science and Technology, Tianjin University, Tianjin, China \\
\email{ \{15112073, yphuang, qzou\}@bjtu.edu.cn, yuhang@email.sc.edu, songwang@cec.sc.edu}
}
\maketitle              
\begin{abstract}
Hazy images are common in real scenarios and many dehazing methods have been developed to automatically remove the haze from images. Typically, the goal of image dehazing is to produce clearer images from which human vision can better identify the object and structural details present in the images. When the ground-truth haze-free image is available for a hazy image, quantitative evaluation of image dehazing is usually based on objective metrics, such as Peak Signal-to-Noise Ratio (PSNR) and Structural Similarity (SSIM). However, in many applications, large-scale images are collected not for visual examination by human. Instead, they are used for many high-level vision tasks, such as automatic classification, recognition and categorization. One fundamental problem here is whether various dehazing methods can produce clearer images that can help improve the performance of the high-level tasks. In this paper, we empirically study this problem in the important task of image classification by using both synthetic and real hazy image datasets. From the experimental results, we find that the existing image-dehazing methods cannot improve much the image-classification performance and sometimes even reduce the image-classification performance.

\end{abstract}
\section{Introduction}

\begin{figure*}[htbp]
	\centering
	\includegraphics[height=2.5cm]{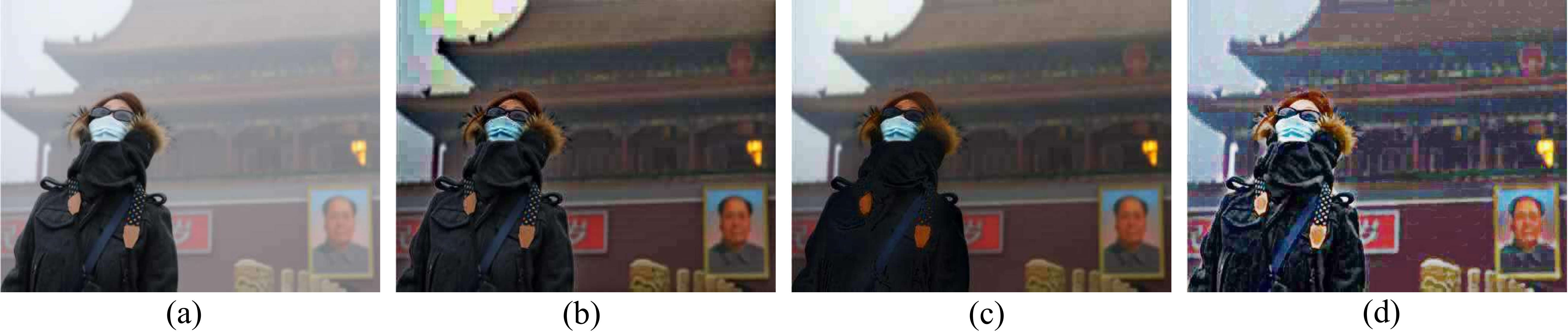}
	\caption{An illustration of image dehazing. (a) A hazy image. (b), (c) and (d) are the images after applying different dehazing methods to the image (a).}
	\label{fig:dehazing}
\end{figure*}

Haze is a very common atmospheric phenomenon where fog, dust, smoke and other particles obscure the clarity of the scene and in practice, many images collected outdoors are contaminated by different levels of
haze, even on a sunny day and in computer vision society, such images are usually called \emph{hazy images}, as shown in Fig.~\ref{fig:dehazing}(a). With intensity blurs and lower contrast, it is usually more difficult to identify object and structural details from hazy images, especially when the level of haze is strong. To address this issue, many \emph{image dehazing} methods~\cite{he2011single,tarel2009fast,tarel2010improved,meng2013efficient,zhu2015fast,berman2016non,cai2016dehazenet,ren2016single,li2017aod} have been developed to remove the haze and try to recover the original clear version of an image. Those dehazing methods mainly rely on various image prior, such as dark channel prior~\cite{he2011single} and color attenuation prior~\cite{zhu2015fast}. As shown in Fig.~\ref{fig:dehazing}, the images after the dehazing are usually more visually pleasing -- it can be easier for the human vision to identify the objects and structures in the image. Meanwhile, many objective metrics, such as Peak Signal-to-Noise Ratio (PSNR)~\cite{huynh2008scope} and Structural Similarity (SSIM)~\cite{wang2004image}, have been proposed to quantitatively evaluate the performance of image dehzaing when the ground-truth haze-free image is available for a hazy image.

However, nowadays large-scale image data are collected not just for visual examination. In many cases, they are collected for high-level vision tasks, such as automatic image classification, recognition and categorization. \emph{One fundamental problem is whether the performance of these high-level vision tasks can be significantly improved if we preprocess all hazy images by applying an image-dehazing method.} On one hand, images after the dehazing are visually clearer with more identifiable details. From this perspective, we might expect the performance improvement of the above vision tasks with image dehazing. On the other hand, most image dehazing methods just process the input images without introducing new information to the images. From this perspective, we may not expect any performance improvement of these vision tasks by using image dehazing since many high-level vision tasks are handled by extracting image information for training classifiers. In this paper, we empirically study this problem in the important task of image classification.

By classifying an image based on its semantic content, \emph{image classification} is an important problem in computer vision and has wide applications in autonomous driving, surveillance and robotics. This problem has been studied for a long time and many well known image databases, such as Caltech-256~\cite{griffin2007caltech}, PASCAL VOCs~\cite{everingham2010pascal} and ImageNet~\cite{deng2009imagenet}, have been constructed for evaluating the performance of image classification. Recently, the accuracy of image classification has been significantly boosted by using deep neural networks. In this paper, we will conduct our empirical study by taking Convolutional Neural Network (CNN), one of the most widely used deep neural networks, as the image classifier and then evaluate the image-classification accuracy with and without the preprocessing of image dehazing. 

More specifically, in this paper we pick eight state-of-the-art image dehazing methods and examine whether they can help improve the image-classification accuracy. To guarantee the comprehensiveness of empirical study, we use both synthetic data of hazy images and real hazy images for experiments and use AlexNet~\cite{krizhevsky2012imagenet}, VGGNet~\cite{simonyan2014very} and ResNet~\cite{he2016deep} for CNN implementation. Note that the goal of this paper is not the development of a new image-dehazing method or a new image-classification method. Instead, we study whether the preprocessing of image dehazing can help improve the accuracy of hazy image classification. We expect this study can provide new insights on how to improve the performance of hazy image classification. 

\section{Related Work}
\label{sec:related work}
Hazy images and their analysis have been studied for many years. Many of the existing researches were focused on developing reliable models and algorithms to remove haze and restore the original clear image underlying an input hazy image. Many models and algorithms have been developed for outdoor image haze removal. For example, in~\cite{he2011single}, dark channel prior was used to remove haze from a single image. In~\cite{meng2013efficient}, an image dehazing method was proposed with a boundary constraint and contextual regularization. In~\cite{zhu2015fast}, color attenuation prior was used for removing haze from a single image. In~\cite{cai2016dehazenet}, an end-to-end method was proposed for removing haze from a single image. In~\cite{ren2016single}, multi-scale convolutional neural networks were used for haze removal. In~\cite{li2017aod}, a haze-removal method was proposed by directly generating the underlying clean image through a light-weight CNN and it can be embedded into other deep models easily. Besides, researchers also investigated haze removal from the images taken at nighttime hazy scenes. For example, in~\cite{li2015nighttime}, a method was developed to remove the nighttime haze with glow and multiple light colors. In~\cite{zhang2017fast},  a fast haze removal method was proposed for nighttime images using the maximum reflectance prior. 

Image classification has attracted extensive attention in the community of computer vision. In the early stage, hand-designed features \cite{yang2009linear} were mainly used for image classification. In recent years, significant progress has been made on image classification, partly due to the creation of large-scale hand-labeled datasets such as ImageNet~\cite{deng2009imagenet}, and the development of deep convolutional neural networks (CNN)~\cite{krizhevsky2012imagenet}. Current state-of-the-art image classification research is focused on training feedforward convolutional neural  networks using ``very deep'' structure~\cite{simonyan2014very,szegedy2015going,he2016deep}. VGGNet~\cite{simonyan2014very}, Inception~\cite{szegedy2015going} and residual learning~\cite{he2016deep} have been proposed to train very deep neural networks, resulting in excellent image-classification performances on clear natural images. In~\cite{liu2015treasure}, a cross-convolutional-layer pooling method was proposed for image classification. In~\cite{wang2016cnn}, CNN is combined with recurrent neural networks (RNN) for improving the performance of image classification. In~\cite{durand2017wildcat}, three important visual recognition tasks, image classification, weakly supervised point-wise object localization and semantic segmentation, were studied in an integrative way. In~\cite{wang2017residual}, a convolutional neural network using attention mechanism was developed for image classification.

Although these CNN-based methods have achieved excellent performance on image classification, most of them were only applied to the classification of clear natural images. Very few of existing works explored the classification of degradation images. In~\cite{agostinelli2013adaptive}, strong classification performance was achieved on corrupted MNIST digits by applying image denoising as an image preprocessing step. In~\cite{tang2012robust}, a model was proposed to recognize faces in the presence of noise and occlusion. In~\cite{wang2016studying}, classification of very low resolution images was studied by using CNN, with applications to face identification, digit recognition and font recognition. In~\cite{jalalvand2016towards}, a preprocessing step of image denoising is shown to be able to improve the performance of image classification under a supervised training framework. In~\cite{chen2016joint}, image denoising and classification were tackled by training a unified single model, resulting in performance improvement on both tasks. Image haze studied in this paper is a special kind of image degradations and, to our best knowledge, there is no systematic study on hazy image classification and whether image dehazing can help hazy image classification. 

\section{Proposed Method}
\label{sec:proposed method}

In this section, we elaborate on the hazy image data, image-dehazing methods, image-classification framework and evaluation metrics used in the empirical study. In the following, we first discuss the construction of both synthetic and real hazy image datasets. We then introduce the eight state-of-the-art image-dehazing methods used in our study. After that, we briefly introduce the CNN-based framework used for image classification. Finally, we discuss the evaluation metrics used in our empirical study.

\subsection{Hazy-Image Datasets}

For this empirical study, we need a large set of hazy images for both image-classifier training and testing. Current large-scale image datasets that are publicly available, such as Caltech-256, PASCAL VOCs and ImageNet, mainly consist of clear images without degradations. In this paper, we use two strategies to get the hazy images. First, we synthesize a large set of 
hazy images by adding haze to clear images using available physical models. Second, we collect a set of real hazy images from the Internet.

\begin{figure}[htbp]
\centering
\includegraphics[height=3.5cm]{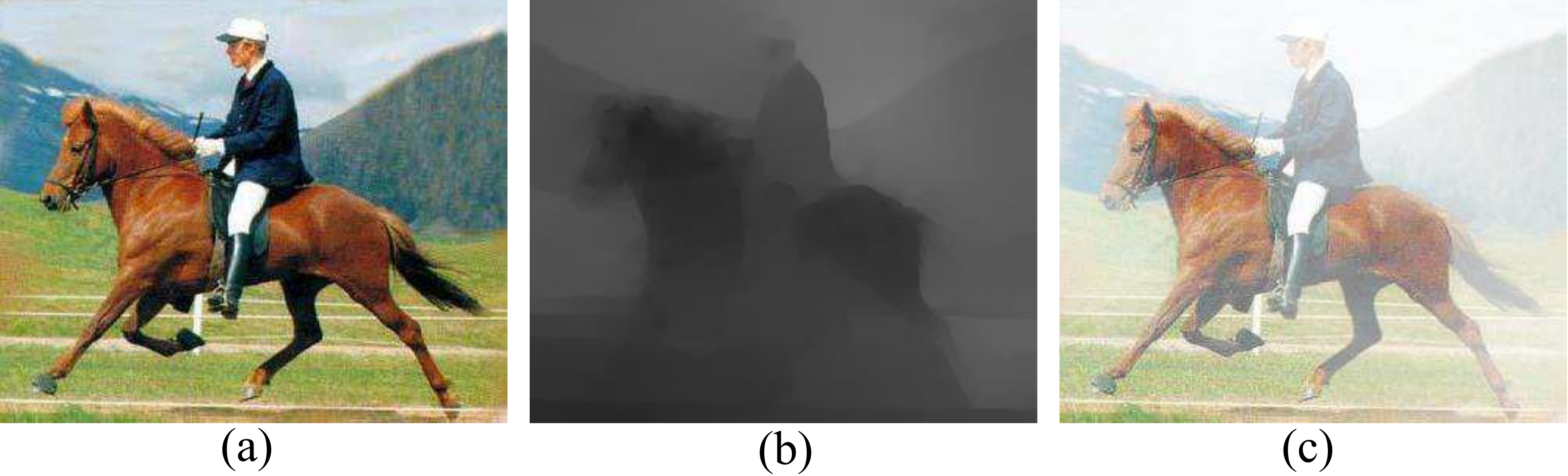}
\caption{An illustration of hazy image synthesis. (a) Clear image. (b) Depth map of (a). (c) Synthetic hazy image.}
\label{fig:synthetic-hazy-image}
\end{figure}

We synthesize hazy images by the following equation~\cite{koschmieder1924theorie}, where the atmospheric scattering model is used to describe the hazy image generation process:
\begin{align}
I(x,y) = t(x,y) \cdot J(x,y) + [1-t(x,y)] \cdot A,
\end{align}
where $(x,y)$ is the pixel coordinate, $I$ is the synthetic hazy image, and $J$ is the original clear image. $A$ is the global atmospheric light. The scene transmission $t(x,y)$ is distance-dependent and defined as
\begin{equation}
t(x,y) = e^{-\beta d(x,y)},
\label{eq:synthesize}
\end{equation}
where $\beta$ is the atmospheric scattering coefficient and $d(x,y)$ is the normalized distance of the scene at pixel $(x,y)$. We compute the depth map $d(x,y)$ of an image by using the algorithm proposed in~\cite{liu2015deep}. An example of such synthetic hazy image, as well as its original clear image and depth map, are shown in Fig.~\ref{fig:synthetic-hazy-image}. In this paper, we take all the images in Caltech-256 to construct synthetic hazy images and the class label of each synthetic image follow the label of the corresponding original clear image. This way, we can use the synthetic images for image classification. 

\begin{figure*}[htbp]
\centering
\includegraphics[height=3.8cm]{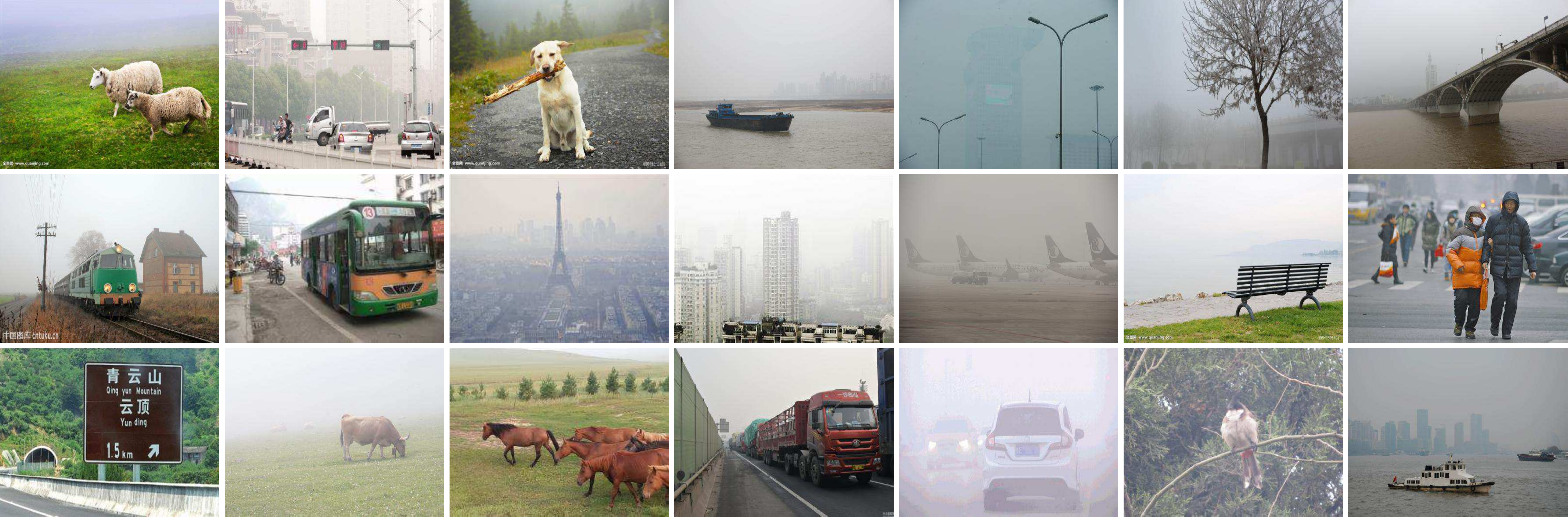}
\caption{Sample hazy images in our new Haze-20 dataset.}
\label{fig:haze-20}
\end{figure*}

While we can construct synthetic hazy images by following well-acknowledged physical models, real haze models can be much more complicated and a study on synthetic hazy image datasets may not completely reflect what we may encounter on real hazy images. To address this issue, we collect a new dataset of hazy images by collecting images from the Internet. This new dataset contains 4,610 images from 20 classes and we named it as \emph{Haze-20}. These 20 image classes are bird (231), boat (236), bridge (233), building (251), bus (222), car (256), chair (213), cow (227), dog (244), horse (237), people (279), plane (235), sheep (204), sign (221), street-lamp (216), tower (230), traffic-light (206), train (207), tree (239) and truck (223), and in the parenthesis is the number of images collected for each class. The number of images per class varies from 204 to 279. Some examples in Haze-20 are shown in Fig.~\ref{fig:haze-20}.  

In this study, we will try the case of training the image-classifier using clear images and testing on hazy images. For synthetic hazy images, we have their original clear images, which can be used for training. For real images in Haze-20, we do not have their underlying clear images. To address this issue, we collect a new \emph{HazeClear-20} image dataset from the Internet, which consists of haze-free images that fall in the same 20 classes as in Haze-20. HazeClear-20 consists of 3,000 images, with 150 images per class.

\subsection{Dehazing Methods}

In this paper we try eight state-of-the-art image-dehazing methods: Dark-Channel Prior (\textbf{DCP})~\cite{he2011single}, Fast Visibility Restoration (\textbf{FVR})~\cite{tarel2009fast}, Improved Visibility(\textbf{IV})~\cite{tarel2010improved}, Boundary Constraint and Contextual Regularization (\textbf{BCCR})~\cite{meng2013efficient}, Color Attenuation Prior (\textbf{CAP})~\cite{zhu2015fast}, Non-local Image Dehazing (\textbf{NLD})~\cite{berman2016non}, DehazeNet (\textbf{DNet})~\cite{cai2016dehazenet}, and \textbf{MSCNN}~\cite{ren2016single}. We examine each of them to see whether it can help improve the performance of hazy image classification.

\begin{itemize}
\item \textbf{DCP} removes haze using dark channel prior, which is based on a key observation -- most local patches of outdoor haze-free images contain some pixels whose intensity is very low in at least one color channel.

\item \textbf{FVR} is a fast haze-removal algorithm based on the median filter. Its main advantage is its fast speed since its complexity is just a linear function of the input-image size.

\item \textbf{IV} enhances the contrast of an input image so that the image visibility is improved. It computes the data cost and smoothness cost for every pixel by using Markov Random Fields.
 
\item \textbf{BCCR} is an efficient regularization method for removing haze. In particular, the inherent boundary constraint on the transmission function combined with a weighted $L_1$-norm based contextual regularization, is modeled into an optimization formulation to recover the unknown scene transmission.

\item \textbf{CAP} removes haze using color attenuation prior that is based on the difference between the saturation and the brightness of the pixels in the hazy image. By creating a linear model, the scene depth of the hazy image is computed with color attenuation prior, where the parameters are learned by a supervised method. 
 
\item \textbf{NLD} is a haze-removal algorithm based on a non-local prior, by assuming that colors of a haze-free image are well approximated by a few hundred of distinct colors in the form of tight clusters in RGB space. In a hazy image, these tight color clusters change due to haze and form lines in RGB space that pass through the airlight coordinate. 
 
\item \textbf{DNet} is an end-to-end haze-removal method based on CNN. The layers of CNN architecture are specially designed to embody the established priors in image dehazing. DNet conceptually consists of four sequential operations -- feature extraction, multi-scale mapping, local extremum and non-linear regression, which are constructed by three convolution layers, a max-pooling, a Maxout unit and a bilinear ReLU activation function, respectively.

\item \textbf{MSCNN} uses a multi-scale deep neural network for image dehazing by learning the mapping between hazy images and their corresponding transmission maps. It consists of a coarse-scale net which predicts a holistic transmission map based on the entire image, and a fine-scale net which refines results locally. The network consists of four operations: convolution, max-pooling, up-sampling and linear combination.
\end{itemize}

\subsection{Image Classification Model}

In this paper, we implement CNN-based model for image classification by using AlexNet~\cite{krizhevsky2012imagenet}, VGGNet-16~\cite{simonyan2014very} and ResNet-50~\cite{he2016deep} on Caffe. The AlexNet~\cite{krizhevsky2012imagenet} has 8 weight layers (5 convolutional layers and 3 fully-connected layers). The VGGNet-16~\cite{simonyan2014very} has 16 weight layers (13 convolutional layers and 3 fully-connected layers). The ResNet-50~\cite{he2016deep} has 50 weight layers (49 convolutional layers and 1 fully-connected layer). For those three networks, the last fully-connected layer has $N$ channels ($N$ is the number of classes).

\subsection{Evaluation Metrics}

We will quantitatively evaluate the performance of image dehazing and the performance of image classification. Other than visual examination, Peak Signal-to-Noise Ratio (PSNR) \cite{huynh2008scope} and Structural Similarity (SSIM) \cite{wang2004image} are widely used for evaluating the performance of image dehazing when the ground-truth haze-free image is available for each hazy image. For image classification, classification accuracy is the most widely used performance evaluation metric. 
 
Note that, both PSNR and SSIM are objective metrics based on image statistics. Previous research has shown that they may not always be consistent with the image-dehazing quality perceived by human vision, which is quite subjective.  In this paper, what we concern about is the performance of image classification after incorporating image dehazing as preprocessing. Therefore, we will study whether PSNR and SSIM metrics show certain correlation to the image classification performance. In this paper, we simply use the classification accuracy $Accuracy=\frac{R}{N}$ to objectively measure the image-classification performance, where $N$ is the total number of testing images and $R$ is the total number of testing images that are correctly classified by using the trained CNN-based models.

\section{Experiments}\label{sec:experiment}

\subsection{Datasets and Experiment Setup}

In this section, we evaluate various image-dehazing methods on the hazy images synthesized from Caltech-256 and our newly collected Haze-20 datasets. 

We synthesize hazy images using all the images in Caltech-256 dataset, which has been widely used for evaluating image classification algorithms. It contains 30,607 images from 257 classes, including 256 object classes and a clutter class. In our experiment, we select six different hazy levels for generating synthetic images. Specifically, we set the parameter $\beta = 0, 1, 2, 3, 4, 5$ respectively in Eq.(\ref{eq:synthesize}) for hazy image synthesis where $\beta = 0$ corresponds to original images in Caltech-256. In Caltech-256, we select 60 images randomly from each class as training images, and the rest are used for testing. Among the training images, 20\% per class are used as a validation set.  We follow this to split the synthetic hazy image data: an image is in training set if it is synthesized from an image in the training set and in testing set otherwise. This way, we have a training set of 60 $\times$ 257 = 15,420 images (60 per class) and a testing set of 30,607 $-$ 15,420 = 15,187 images for each hazy level.

For the collected real hazy images in Haze-20, we select 100 images randomly from each class as training images, and the rest are used for testing. Among the training images, 20\% per class are used as a validation set. So, we have a training set of $100 \times 20 = 2,000$ images and a testing set of $4,610-2,000 = 2,610$ images. For HazeClear-20 dataset, we also select 100 images randomly from each class as training images, and the rest are used for testing. Among the training images, 20\% per class are used as a validation set. So, we have a training set of $100 \times 20 = 2,000$ images and a testing set of $50\times 20 = 1,000$ images.

While the proposed CNN model can use AlexNet, VGGNet, ResNet or another network structures, for simplicity, we use AlexNet, VGGNet-16, ResNet-50 on Caffe in this paper. The CNN architectures are pre-trained on ImageNet dataset that consists of 1,000 classes with 1.2 million training images. We then use the collected images to fine-tune the pre-trained model for image classification, in which we change the number of channels in the last fully connected layer from 1,000 to $N$, where $N$ is the number of classes in our datasets. To more comprehensively explore the effect of haze-removal to image classification, we study different combinations of the training and testing data, including training and testing on images without applying image dehazing, training and testing on images after dehazing, and training on clear images but testing on hazy images.

\subsection{Quantitative Comparisons on Synthetic and Real Hazy Images}

\begin{figure*}[htbp]
\centering
\includegraphics[height=4.5cm]{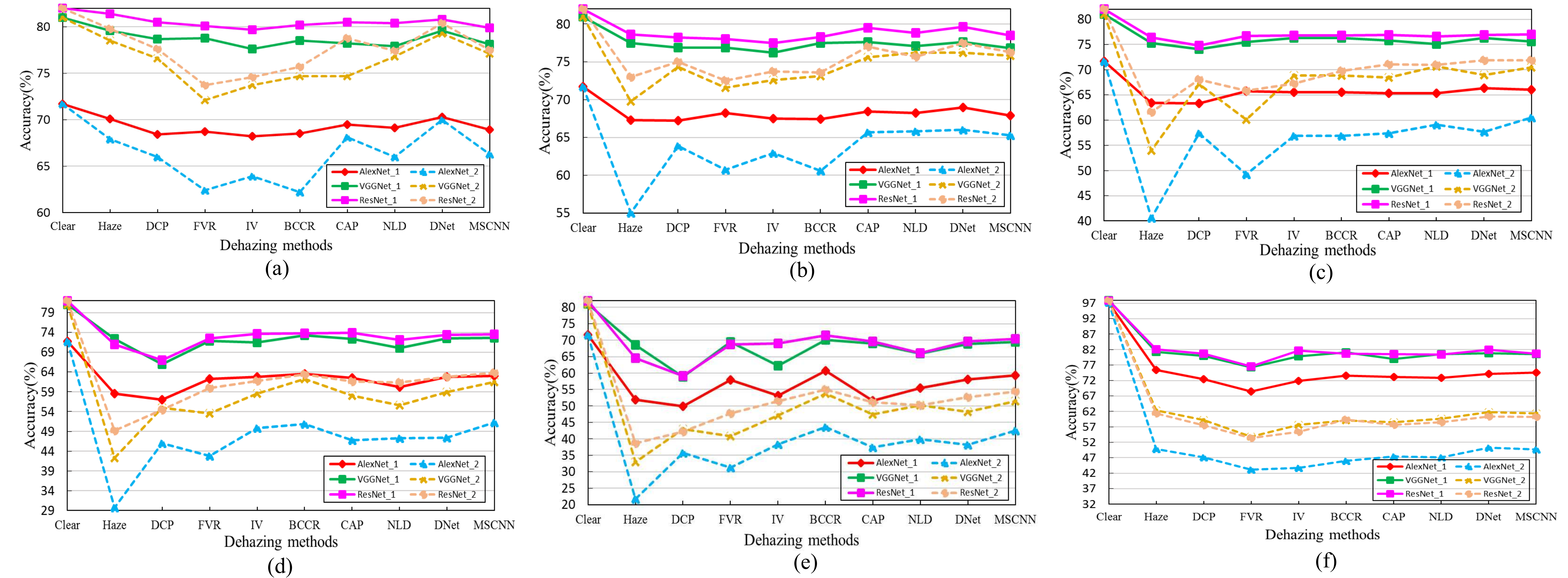}
\caption{The classification accuracy on different hazy images. (a-e) Classification accuracies on testing synthetic hazy images with $\beta=1, 2, 3, 4, 5$, respectively. (f) Classification accuracy on the testing images in Haze-20.} 
\label{fig:caltech256_haze20_accuracy}
\end{figure*}

To verify whether haze-removal preprocessing can improve the performance of hazy image classification, we test on the  synthetic and real hazy images with and without haze removal for quantitative evaluation. The classification results are shown in Fig.~\ref{fig:caltech256_haze20_accuracy}, where (a-e) are the classification accuracies on testing synthetic hazy images with $\beta=1, 2, 3, 4, 5$, respectively using different dehazing methods. For these five curve figures, the horizontal axis lists different dehazing methods, where ``Clear" indicates the use of the testing images in the original Caltech-256 datasets and this assumes a perfect image dehazing in the ideal case. The case of ``Haze" indicates the testing on the hazy images without any dehazing. (f) is the classification accuracy on the testing images in Haze-20 using different dehazing methods, where ``Clear" indicates the use of testing images in HazeClear-20 and ``Haze" indicates the use of testing images in Haze-20 without any dehazing. \emph{AlexNet\_1}, \emph{VGGNet\_1} and \emph{ResNet\_1} represent the case of training and testing on the same kinds of images, e.g., training on the training images in Haze-20 after DCP dehazing, then testing on testing images in Haze-20 after DCP dehazing, by using AlexNet, VGGNet and ResNet, respectively. \emph{AlexNet\_2}, \emph{VGGNet\_2} and \emph{ResNet\_2} represent the case of training on clear images, i.e., for (a-e), we train on training images in original Caltech-256, and for (f), we train on training images in HazeClear-20, by using AlexNet, VGGNet and ResNet, respectively. 

We can see that when we train CNN models on clear images and test them on hazy images with and without haze removal (e.g., \emph{AlexNet\_2}, \emph{VGGNet\_2} and \emph{ResNet\_2}), the classification performance drop significantly. From Fig.~\ref{fig:caltech256_haze20_accuracy}(e), image classification accuracy drop from 71.7\% to 21.7\% when images have a haze level of  $\beta=5$ by using AlexNet. Along the same curve shown in Fig.~\ref{fig:caltech256_haze20_accuracy}(e),  we can see that by applying a dehazing method on the testing images, the classification accuracy can move up to 42.5\% (using MSCNN dehazing). But it is still much lower than 71.7\%, the accuracy on classifying original clear images. These experiments indicate that haze significantly affects the accuracy of CNN-based image classification when training on original clear images. However, if we directly train the classifiers on the hazy image of the same level, the classification accuracy moves up to 51.9\%, as shown in the red curve in Fig.~\ref{fig:caltech256_haze20_accuracy}(e), where no dehazing is involved in training and testing images. Another choice is to apply the same dehazing methods to both training and testing images: From results shown in all the six subfigures in Fig.~\ref{fig:caltech256_haze20_accuracy}, we can see that the resulting accuracy is similar to the case where no dehazing is applied to training and testing images. This indicates that the dehazing conducted in this study does not help image classification. We believe this is due to the fact that the dehazing does not introduce new information to the image.

\begin{figure}[htbp]
\centering
\includegraphics[height=4.1cm]{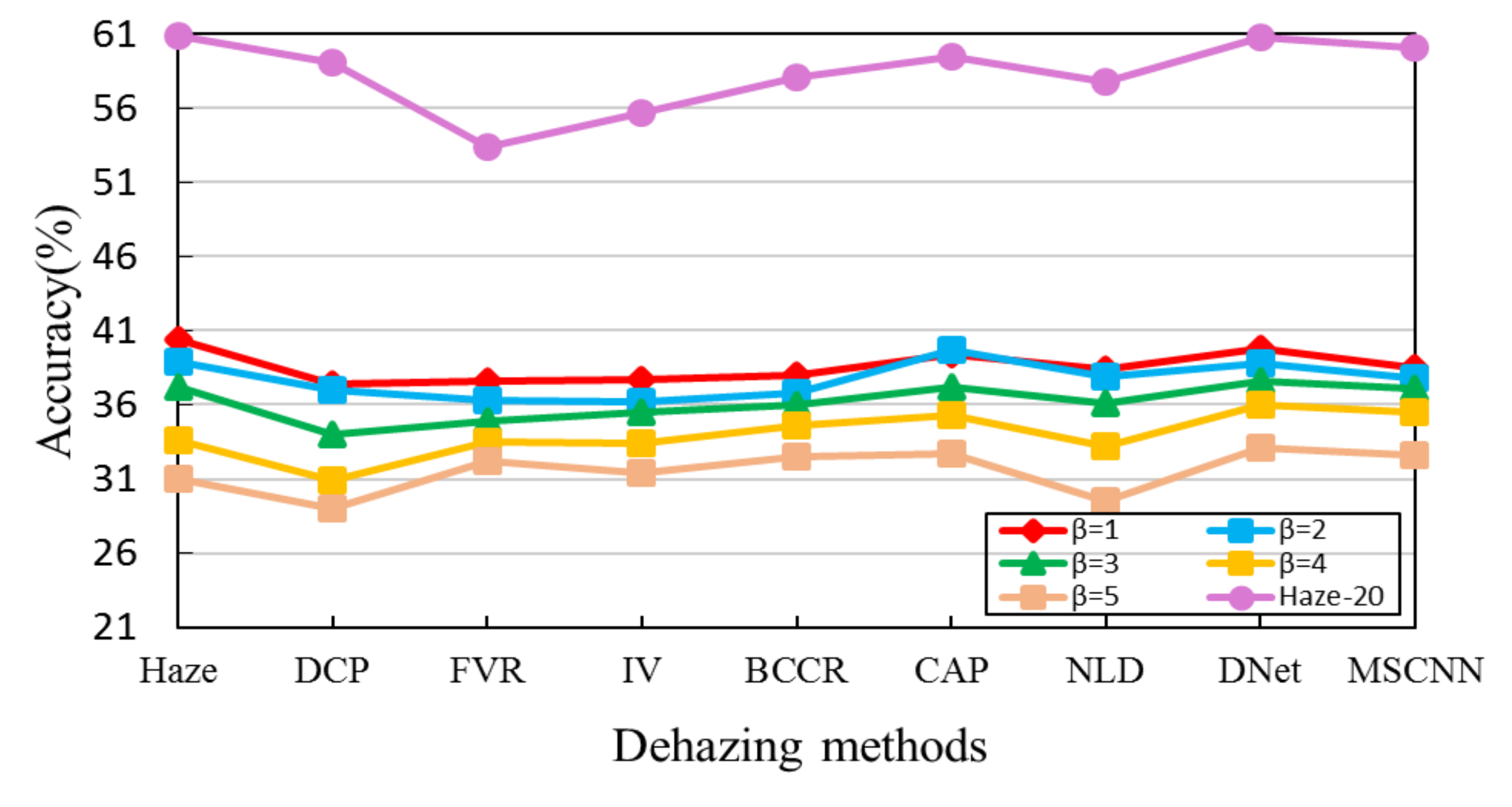}
\caption{Classification accuracy (\%)  on synthetic and real-world hazy images by using a non-CNN-based image classification method. Here the same kinds of images are used for training, i.e., building the basis for sparse coding, and testing, just like the case corresponding to the solid curves (\emph{AlexNet\_1}, \emph{VGGNet\_1} and \emph{ResNet\_1} ) in Fig.~\ref{fig:caltech256_haze20_accuracy}. } 
\label{fig:scspm_accuracy}
\end{figure}

There are also many non-CNN-based image classification methods. While it is difficult to include all of them into our empirical study, we try the one based on sparse coding~\cite{yang2009linear} and the results are shown in Fig.~\ref{fig:scspm_accuracy}, where $\beta=1,2,3,4,5$ represent haze levels of synthetic hazy images in Caltech-256 dataset and \emph{Haze-20} represents Haze-20 dataset. For this specific non-CNN-based image classification method, we can get the similar conclusion that the tried dehazing does not help image classification, as shown in Fig.~\ref{fig:scspm_accuracy}. Comparing Figs.~\ref{fig:caltech256_haze20_accuracy} and \ref{fig:scspm_accuracy}, we can see that the classification accuracy of this non-CNN-based method is much lower than the state-of-the-art CNN-based methods. Therefore, we focus on CNN-based image classification in this paper.

\subsection{Training on Mixed-Level Hazy Images}

\begin{figure*}[!t]
\centering
\includegraphics[height=4.5cm]{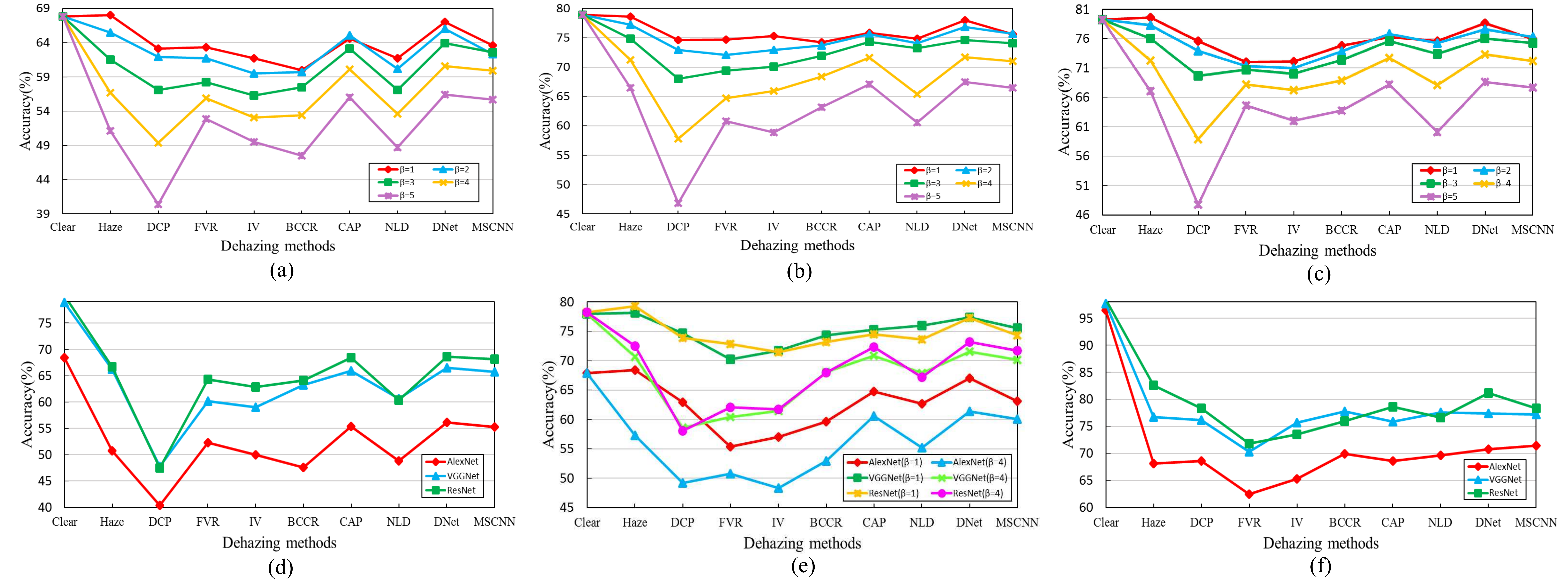}
\caption{Classification accuracy when training on mixed-level hazy images. (a, b, c) Mix all six levels of synthetic images. (d) Mix two levels $\beta=0$ and $\beta=5$. (e) Mix two levels $\beta=1$ and $\beta=4$. (f) Mix Haze-20 and HazeClear-20.} 
\label{fig:caltech256-haze20-mix}
\end{figure*}

For more comprehensive analysis of dehazing methods, we conduct experiments of training on hazy images with mixed haze levels. For synthetic dataset, we try two cases. In Case 1, we mix all six levels of hazy images by selecting 10 images per class from each level of hazy images as training set and among the training images, two images per class per haze level are taken as validation set. We then test on the testing images of the involved haze levels -- actually all six levels for this case -- respectively. Results are shown in Fig.~\ref{fig:caltech256-haze20-mix}(a), (b) and (c) when using AlexNet, VGGNet and ResNet respectively.  In Case 2, we randomly choose images from two different haze levels and mix them. In this case, 30 images per class per level are taken as training images and among the training images, 6 images per class per level are used as validation images. This way we have 60 images per class for training. Similarly, we then test on the testing images of the involved two haze levels, respectively. Results are shown in Fig.~\ref{fig:caltech256-haze20-mix}(d) and (e)
for four different kinds of level combinations, respectively. For real hazy images, we mix clear images in HazeClear-20 and hazy images in Haze-20 by picking 50 images per class for training and then test on the testing images in Haze-20 and HazeClear-20 respectively. Results are shown in Fig.~\ref{fig:caltech256-haze20-mix}(f). Similarly, combining all the results, the use of dehazing does not clearly improve the image classification accuracy, over the case of directly training and testing on hazy images.

\subsection{Performance Evaluation of Dehazing Methods}

\begin{figure*}[htbp]
\centering
\includegraphics[height=3.1cm]{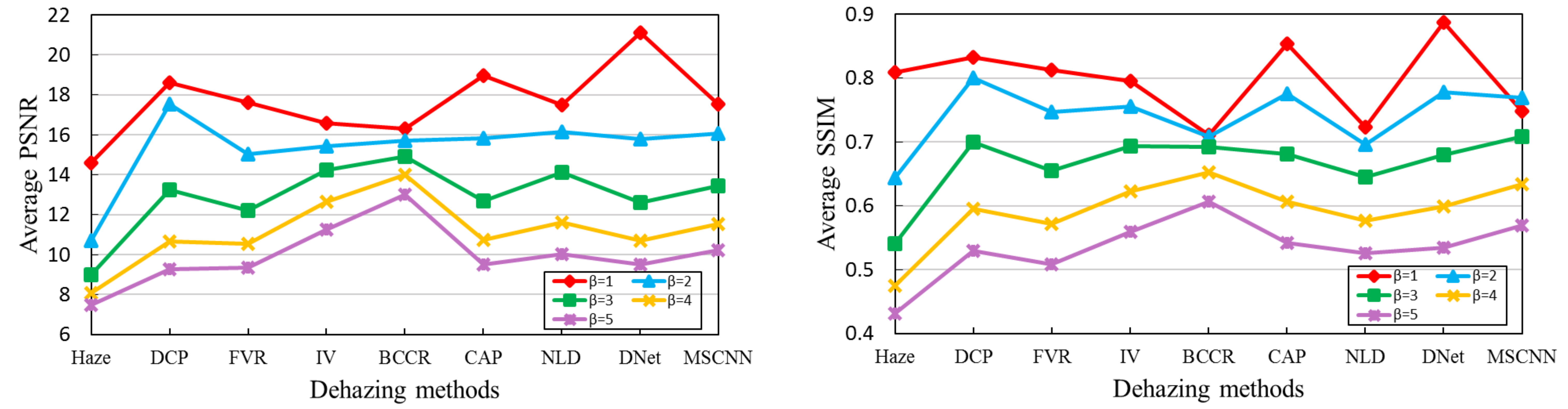}
\caption{Average PSNR and SSIM values on synthetic image dataset at different haze levels.} 
\label{fig:psnr_ssim}
\end{figure*}

In this section, we study whether there is a correlation between the dehazing metrics PSNR/SSIM and the image classification performance. On the synthetic images, we can compute the metrics PSNR and SSIM on all the dehazing results, which are shown in Fig.~\ref{fig:psnr_ssim}. In this figure, the PSNR and SSIM values are averaged over the respective testing images. We pick the red curves (\emph{AlexNet\_1}) from Fig.~\ref{fig:caltech256_haze20_accuracy}(a-e) and for each haze level in $\beta=1, 2, 3, 4, 5$, we rank all the dehazing methods based on the classification accuracy. We then rank these methods based on average PSNR and SSIM at the same haze level. Finally we calculate the rank correlation between image classification and PSNR/SSIM at each haze level. Results are shown in Table~\ref{table:rank-correlation}. Negative values indicate negative correlation, positive values indicate positive correlation and the greater the absolute value, the higher the correlation. We can see that their correlations are actually low, especially when $\beta=3$.

\setlength{\tabcolsep}{5pt}
\renewcommand\arraystretch{1.5}
\begin{table}
\begin{center}
\caption{The rank correlation between image-classification accuracy and PSNR/SSIM at each haze level.}
\label{table:rank-correlation}
\begin{tabular}{|l|l|l|l|l|l|}
\hline
Correlation &$\beta=1$ &$\beta=2$ &$\beta=3$ &$\beta=4$ &$\beta=5$\\
\hline
(Accuracy, PSNR) &-0.3095 &0.3571 &0.0952 &-0.2143 &0.1905\\
\hline
(Accuracy, SSIM) &-0.2381 &-0.5238 &-0.0714 &0.6905 &0.6190\\
\hline
\end{tabular}
\end{center}
\end{table}
\setlength{\tabcolsep}{1.4pt}

\subsection{Subjective Evaluation}

\begin{figure}[htbp]
\centering
\includegraphics[height=4.2cm]{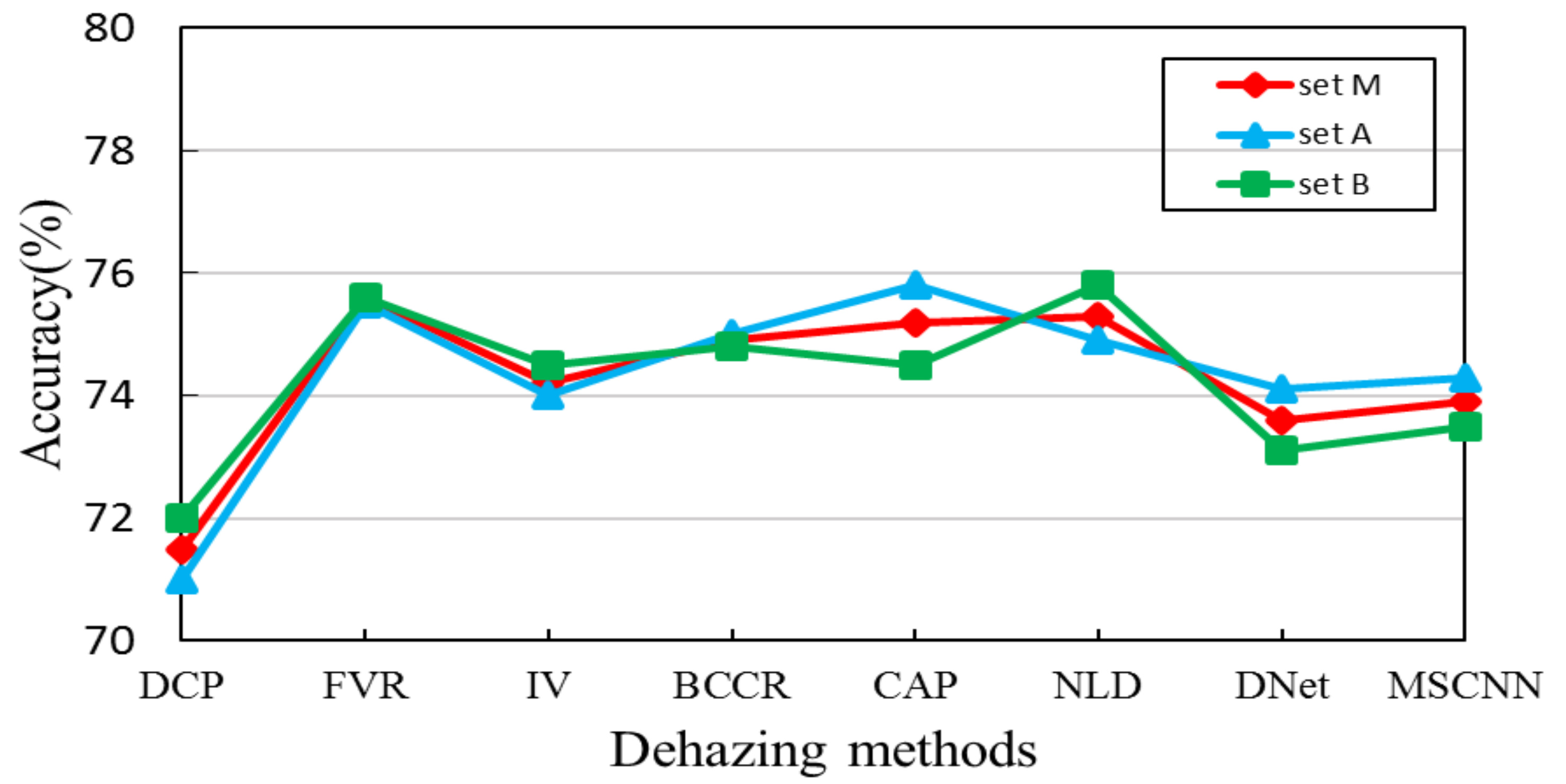}
\caption{Classification accuracy of different sets of dehazed images subjectively selected by human.} 
\label{fig:subjective_vggnet}
\end{figure}

In this section, we conduct an experiment for subjective evaluation of the image dehazing. By observing the dehazed images, we randomly select 10 images per class with $\beta=3$ and subjectively divide them into 5 with better dehazing effect and 5 with worse dehazing effect. This way, we have 2,570 images in total (set M) and 1,285 images each with better dehazing (set A) and worse dehazing (set B). Classification accuracy (\%) using VGGNet is shown in Fig.~\ref{fig:subjective_vggnet} and we can see that there is no significant accuracy difference for these three sets. This indicates that the classification accuracy is not consistent with the human subjective evaluation of the image dehazing quality.

\subsection{Feature Reconstruction}

\begin{figure*}[htbp]
\centering
\includegraphics[height=2.9cm]{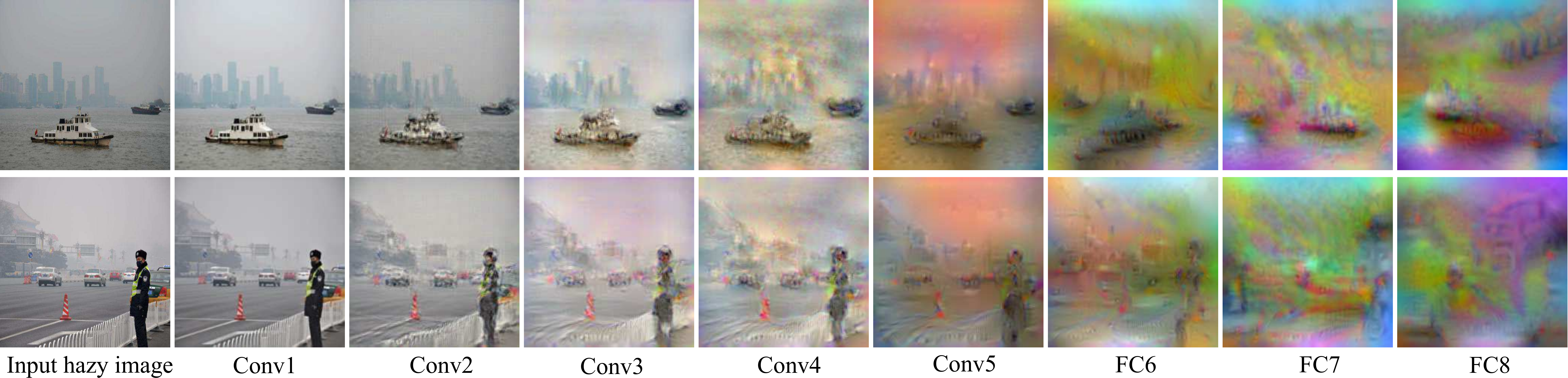}
\caption{Sample feature reconstruction results for two images, shown in two rows respectively. The leftmost column shows the input hazy images and the following columns are the images reconstructed from different layers in AlexNet.} 
\label{fig:reconstruction}
\end{figure*}

The CNN networks used for image classification consists of multiple layers to extract deep image features. One interesting question is whether certain layers in the trained CNN actually perform image dehazing implicitly. We picked a reconstruction method~\cite{mahendran2015understanding} to reconstruct the image according to feature maps of all the layers in AlexNet. The reconstruction results are shown in Fig.~\ref{fig:reconstruction},  from which we can see that, for the first several layers, the reconstructed images do not show any dehazing effect. For the last several layers, the reconstructed images have been distorted, let alone dehazing. One possibility of this is that many existing image dehazing methods aim to please human vision system, which may not be good to CNN-based image classification. Meanwhile, many existing image dehazing methods introduce information loss, such as color distortion, and may increase the difficulty of image classification.

\subsection{Feature Visualization}
\begin{figure*}[htbp]
\centering
\includegraphics[height=5.7cm]{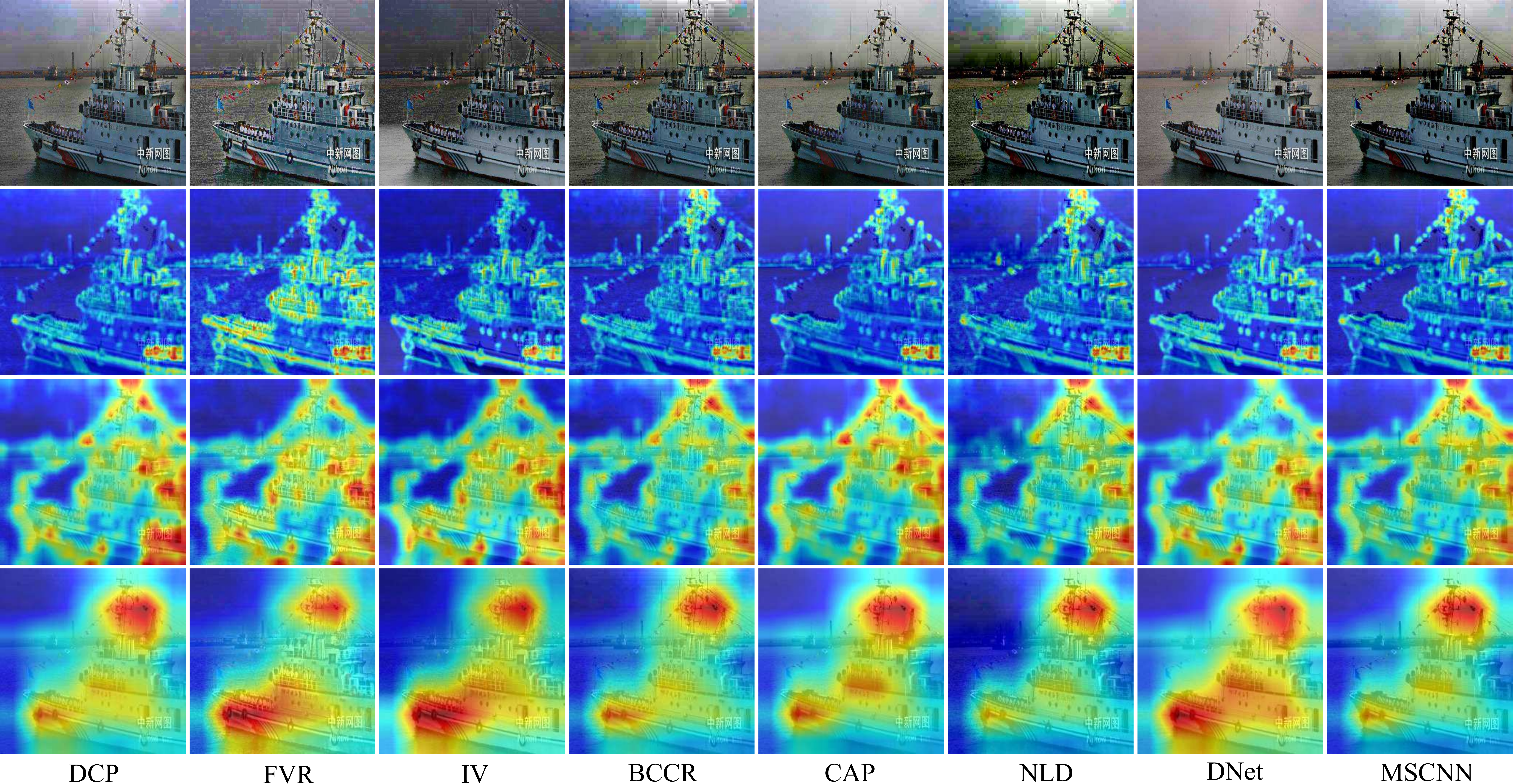}
\caption{Activations of hidden layers of VGGNet on image classification. From top to bottom are the haze-removal images, and the activations at $pool_1$, $pool_3$ and $pool_5$ layers, respectively.} 
\label{fig:visual}
\end{figure*}

In order to further analyze different dehazing methods, we extract and visualize the features at hidden layers using VGGNet. For an input image with size $H \times W$, the activations of a convolution layer is formulated as an order-3 tensor with $H \times W \times D$ elements, where $D$ is the number of channels. The term ``activations" is a feature map of all the channels in a convolution layer. The activations in haze-removal images with different dehazing methods are displayed in Fig.~\ref{fig:visual}. From top to bottom are haze-removal images, and the activations at $pool_1$, $pool_3$ and $pool_5$ layers, respectively. We can see that different dehazing methods actually have different activations, such as the activations of $pool_5$ layer of NLD and DNet.

\section{Conclusions}

In this paper, we conducted an empirical study to explore the effect of image dehazing to the performance of CNN-based image classification on synthetic and real hazy images. We used physical haze models to synthesize a large number of hazy images with different haze levels for training and testing. We also collected a new dataset of real hazy images from the Internet and it contains 4,610 images from 20 classes. We picked eight well-known dehazing methods for our empirical study. Experimental results on both synthetic and real hazy datasets show that the existing dehazing algorithms do not bring much benefit to improve  the CNN-based image-classification accuracy, when compared to the case of directly training and testing on hazy images. Besides, we analyzed the current dehazing evaluation measures based on pixel-wise errors and local structural similarities and showed that there is not much correlation between these dehazing metrics and the image-classification accuracy when the images are preprocessed by the exsiting dehazing methods. While we believe this is due to the fact that image dehazing does not introduce new information to help image classification, we do not exclude the possibility that the existing image-dehazing methods are not sufficiently good in recovering the original clear image and better image-dehazing methods developed in the future may help improve image classification. We hope this study can draw more interests from the community to work on the important problem of haze image classification, which plays a critical role in applications such as autonomous driving, surveillance and robotics.


%
%
%
\clearpage
\bibliographystyle{splncs04}
\bibliography{egbib}
  
\end{document}